%% file: iclr2023_conference_tinypaper.tex
\documentclass{article} 
\usepackage{iclr2023_conference_tinypaper,times}

\input{math_commands.tex}

\usepackage{algorithm}
\usepackage{algpseudocode}
\usepackage[hidelinks]{hyperref}

\usepackage{url}

\usepackage{multirow}
\usepackage{graphicx}
\usepackage{subcaption}
\usepackage{wrapfig}

\title{L-TUNING: Synchronized Label Tuning for Prompt and Prefix in LLMs}

\author{Md Kowsher$^{1}$, Md. Shohanur Islam Sobuj$^{2}$, Asif Mahmud$^{3}$, \bf Nusrat Jahan Prottasha$^{1}$, Prakash Bhat$^{4}$ \vspace{0cm}\\
$^1$Stevens Institute of Technology, $^2$Hajee Mohammad Danesh Science \& Technology University\\
$^3$Noakhali Science \& Technology University, $^4$Amazon \\
}

\iclrfinalcopy 
\begin{document}

\maketitle

\begin{abstract}
Efficiently fine-tuning Large Language Models (LLMs) for specific tasks presents a considerable challenge in natural language processing. Traditional methods, like prompt or prefix tuning, typically rely on arbitrary tokens for training, leading to prolonged training times and generalized token use across various class labels. To address these issues, this paper introduces L-Tuning, an efficient fine-tuning approach designed for classification tasks within the Natural Language Inference (NLI) framework. Diverging from conventional methods, L-Tuning focuses on the fine-tuning of label tokens processed through a pre-trained LLM, thereby harnessing its pre-existing semantic knowledge. This technique not only improves the fine-tuning accuracy and efficiency but also facilitates the generation of distinct label embeddings for each class, enhancing the model's training nuance. Our experimental results indicate a significant improvement in training efficiency and classification accuracy with L-Tuning compared to traditional approaches, marking a promising advancement in fine-tuning LLMs for complex language tasks. \\
Code is available at: \textcolor{red}{\href{https://github.com/Kowsher/L-Tuning}{\texttt{https://github.com/Kowsher/L-Tuning}}}.

\end{abstract}
  
  
\vspace{-6pt}
\section{Introduction}
\vspace{-4pt}
The advent of LLM has marked a significant milestone in NLP \citep{ge2023openagi}. However, the effective utilization of LLMs often depends on fine-tuning techniques such as prompt or prefix tuning \citep{peng2023soft}. Traditional methods, which typically involve training arbitrary tokens for all labels to guide the model, encounter limitations in the context of LLMs \citep{liu2022p, lester2021power, gu2021ppt, han2022ptr}.  Due to the non-semantic nature of these tokens, requiring extensive training for effective integration. Additionally, the use of identical tokens across all classes leads to suboptimal performance due to the lack of semantic differentiation among the classes.
\vspace{-15pt}
\begin{wrapfigure}{r}{0.7\textwidth}

    \centering
    \includegraphics[width=\linewidth]{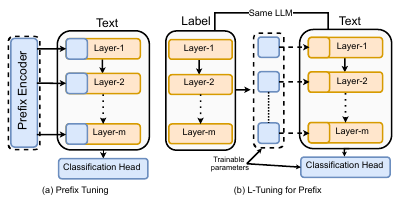}
    \captionsetup{skip=-2pt} 
    \caption{Prefix tuning and L-Tuning for Prefix, highlighting label embedding integration and classification pathways.}
    \label{fig:a}
\vspace{-15pt}
\end{wrapfigure}

To surmount these challenges, we introduce L-Tuning, an innovative approach to prompt and prefix tuning, particularly tailored for classification tasks within the NLI framework \citep{kowsher2023contrastive}. Distinct from traditional methods, L-Tuning leverages label tokens that are initially processed through the pre-trained LLM. This strategy effectively utilizes the LLM's inherent semantic knowledge, enabling more efficient and precise optimization. Furthermore, L-Tuning employs unique label tokens for each class, thereby providing a more refined method for fine-tuning. Empirical evidence suggests that L-Tuning significantly outperforms conventional prompt and prefix tuning in LLMs, both in terms of reducing training time and enhancing performance in classification tasks.
\section{L-Tuning Procedure}
\label{sec:method}
Consider a classification task with \(K\) distinct classes. Let our training dataset be denoted by \( \mathcal{D} = \{(\mathbf{x}_i, y_i)\}_{i=1}^N \), where \( \mathbf{x}_i \) represents the input text and \( y_i \) is the true label corresponding to \( \mathbf{x}_i \). Our objective is to fine-tune a pre-trained LLM \(\mathcal{M}\) for the classification task within an NLI framework; while keeping the majority of the LLM parameters \(\Theta\) frozen. Each input instance to our system is a pair consisting of a text input \(\mathbf{x}_i\) and a label input \(\mathbf{y}_i\), reflecting an NLI setting. The goal is to ascertain the veracity of \( y_i \) as the correct label for \( \mathbf{x}_i \).

\textbf{L-Tuning for Prefix:}
In contrast to traditional prefix tuning, we utilize an $m$'s layers pre-trained LLM \(\mathcal{M}\) with parameters frozen to obtain prefix embeddings directly from the label tokens. For a label token sequence \(\mathbf{y}_i\) with length \(l\) and model dimension \(d\), we derive its hidden representation \(\mathbf{h}_i = \mathcal{M}_{\Theta_{\text{frozen}}}(\mathbf{y}_i) \in \mathbb{R}^{l \times d}\). Given that \(\mathbf{h}_i\) is a matrix, we apply a self-attention pooling function \(\mathcal{F}\), parameterized by \(\Phi\), to transform it into a suitable form. A subsequent transformation function \(Z\), with parameters \(\Psi\), is then used to generate the layer's hidden states of \(\mathcal{M}\): \(\mathbf{p}_i = Z_{\Psi}(\mathcal{F}_{\Phi}(\mathbf{h}_i)) \in \mathbf{R}^{ m \times l \times d}\). These embeddings are used by the classification head \(\mathcal{C}\), parameterized by \(\zeta\), in conjunction with the text input \(\mathbf{x}_i\) to produce the final output \(\mathbf{o}_i = \mathcal{C}_{\zeta}(\mathcal{M}_{\Theta_{\text{frozen}}}(\mathbf{x}_i, \mathbf{p}_i))\). Our training objective minimizes the loss function \(\mathcal{L}\), which assesses the discrepancy between \(\mathbf{o}_i\) and the binary target label \(\mathbf{c} \in \{0,1\}\), indicating whether \(\mathbf{y}_i\) is the correct label for \(\mathbf{x}_i\).
\begin{equation}
    \min_{\Phi, \Psi, \zeta} \mathcal{L}\left(\mathcal{C}_{\zeta}(\mathcal{M}_{\Theta_{\text{frozen}}}(\mathbf{x_i}, Z_{\Psi}(\mathcal{F}_{\Phi}(\mathcal{M}_{\Theta_{\text{frozen}}}(\mathbf{y_i}))))), \mathbf{c}\right).
\end{equation}
This method allows the fine-tuning process to focus specifically on the representation and understanding of labels, leveraging the intrinsic knowledge encapsulated in \(\Theta\) while refining the model's ability to map textual inputs to their corresponding labels through adjustments to \(\Phi\), \(\Psi\) and \(\zeta\) alone.

\textbf{L-Tuning for Prompt:} For the prompt, we acquire label embedding \(\mathbf{e(y_i)} = \mathcal{G}_{\gamma}(\mathbf{h_i}) \in \mathbb{R}^{l \times d}\), where \(\mathcal{G}\) is a trainable transformation function, parameterized by \(\gamma\), to generate label embeddings. We also derive text data embeddings from the frozen LLM embedding as \(\mathbf{e(x_i)} \in \mathbb{R}^{n \times d}\), where \(n\) is the sequence length of the text sample \(\mathbf{x_i}\). The classification head \(\mathcal{C}\) is then defined as the concatenation of both embeddings: $\mathbf{o}_i = \mathcal{C}_{\zeta}(\mathcal{M}_{\Theta_{\text{frozen}}}(\mathbf{e(y_i)} \oplus \mathbf{e(x_i)}))$. The training objective for L-Tuning for prompt can be defined as:
\begin{equation}
    \min_{\gamma, \zeta} \mathcal{L}\left(\mathcal{C}_{\zeta}(\mathcal{M}_{\Theta_{\text{frozen}}}(\mathcal{G}_{\gamma}(\mathcal{M}_{\Theta_{\text{frozen}}}(\mathbf{e(y_i)}))\oplus \mathbf{e(x_i)})), \mathbf{c}\right).
\end{equation}
\section{Experiments \& Conclusion}

\begin{table}[ht]
\caption{A comparative analysis of pre-trained LMs and LLMs-7B. The comparison is conducted across various input methodologies — Prefix, Prompt, LT-prefix, and LT-prompt. Highlighted within are the performance metrics, specifically accuracy scores, for each combination of model and dataset, illustrating the efficacy of each input methodology in enhancing model performance.}
\vspace{-7pt}
\label{tab:performance}
\resizebox{\columnwidth}{!}{%
\begin{tabular}{c|ccc|ccc|ccc}
\hline
\textbf{Dataset} &
  \multicolumn{3}{c|}{\textbf{Cola}} &
  \multicolumn{3}{c|}{\textbf{RTE}} &
  \multicolumn{3}{c}{\textbf{sst-2}} \\ \hline
 \textbf{LMs}&
  \multicolumn{1}{c|}{\textbf{BERT}} &
  \multicolumn{1}{c|}{\textbf{RoBERTa}} &
  \textbf{DeBERTa} &
  \multicolumn{1}{c|}{\textbf{BERT}} &
  \multicolumn{1}{c|}{\textbf{RoBERTa}} &
  \textbf{DeBERTa} &
  \multicolumn{1}{c|}{\textbf{BERT}} &
  \multicolumn{1}{c|}{\textbf{RoBERTa}} &
  \textbf{DeBERTa} \\ \hline
{Prefix} &
  \multicolumn{1}{c|}{0.807} &
  \multicolumn{1}{c|}{0.798} &
  0.834 &
  \multicolumn{1}{c|}{0.682} &
  \multicolumn{1}{c|}{0.635} &
  0.711 &
  \multicolumn{1}{c|}{0.912} &
  \multicolumn{1}{c|}{0.940} &
  0.938 \\ \hline
{LT-Prefix} &
  \multicolumn{1}{c|}{0.812} &
  \multicolumn{1}{c|}{0.803} &
  0.840 &
  \multicolumn{1}{c|}{0.701} &
  \multicolumn{1}{c|}{0.651} &
  0.729 &
  \multicolumn{1}{c|}{0.920} &
  \multicolumn{1}{c|}{0.942} &
  0.947 \\ \hline
{Prompt} &
  \multicolumn{1}{c|}{0.791} &
  \multicolumn{1}{c|}{0.784} &
  0.812 &
  \multicolumn{1}{c|}{0.661} &
  \multicolumn{1}{c|}{0.594} &
  0.672 &
  \multicolumn{1}{c|}{0.891} &
  \multicolumn{1}{c|}{0.931} &
  0.907 \\ \hline
{LT-Prompt} &
  \multicolumn{1}{c|}{0.802} &
  \multicolumn{1}{c|}{0.791} &
  0.819 &
  \multicolumn{1}{c|}{0.682} &
  \multicolumn{1}{c|}{0.604} &
  0.679 &
  \multicolumn{1}{c|}{0.902} &
  \multicolumn{1}{c|}{0.939} &
  0.927 \\ \hline
\textbf{LLMs} &
  \multicolumn{1}{c|}{\textbf{Falcon}} &
  \multicolumn{1}{c|}{\textbf{Bloom}} &
  \textbf{Llama-2} &
  \multicolumn{1}{c|}{\textbf{Falcon}} &
  \multicolumn{1}{c|}{\textbf{Bloom}} &
  \textbf{Llama-2} &
  \multicolumn{1}{c|}{\textbf{Falcon}} &
  \multicolumn{1}{c|}{\textbf{Bloom}} &
  \textbf{Llama-2} \\ \hline
{Prefix} &
  \multicolumn{1}{c|}{0.799} &
  \multicolumn{1}{c|}{0.824} &
  0.811 &
  \multicolumn{1}{c|}{0.652} &
  \multicolumn{1}{c|}{0.634} &
  0.692 &
  \multicolumn{1}{c|}{0.848} &
  \multicolumn{1}{c|}{0.857} &
  0.881 \\ \hline
{LT-prefix} &
  \multicolumn{1}{c|}{0.823} &
  \multicolumn{1}{c|}{0.842} &
  0.852 &
  \multicolumn{1}{c|}{0.672} &
  \multicolumn{1}{c|}{0.684} &
  0.731 &
  \multicolumn{1}{c|}{0.901} &
  \multicolumn{1}{c|}{0.909} &
  0.941 \\ \hline
{Prompt} &
  \multicolumn{1}{c|}{0.772} &
  \multicolumn{1}{c|}{0.835} &
  0.793 &
  \multicolumn{1}{c|}{0.607} &
  \multicolumn{1}{c|}{0.615} &
  0.662 &
  \multicolumn{1}{c|}{0.804} &
  \multicolumn{1}{c|}{0.825} &
  0.845 \\ \hline
{LT-prompt} &
  \multicolumn{1}{c|}{0.816} &
  \multicolumn{1}{c|}{0.817} &
  0.821 &
  \multicolumn{1}{c|}{0.638} &
  \multicolumn{1}{c|}{0.660} &
  0.691 &
  \multicolumn{1}{c|}{0.873} &
  \multicolumn{1}{c|}{0.882} &
  0.911 \\ \hline
\end{tabular}%
}
\end{table}

In our comparative study of various language models, including BERT \citep{devlin-etal-2019-bert}, RoBERTa \citep{liu2019roberta}, DeBERTa \citep{he2021deberta}, Falcon \citep{penedo2023refinedweb}, Bloom \citep{workshop2023bloom}, and Llama-2 \citep{touvron2023llama}, we observed distinct performance enhancements across Cola, RTE, and sst-2 datasets \citep{wang2019glue} using LT-prefix and LT-prompt tuning methods. Notably, L-Tuning demonstrated a modest improvement of 0-2\% for standard language models (LMs) like BERT and RoBERTa, but its impact was more pronounced in large language models (LLMs) like Bloom and Llama-2, showing improvements of 2-6\%. This indicates that L-Tuning's efficacy is particularly significant in the context of LLMs, underscoring its potential as a scalable and efficient approach to optimizing advanced language processing systems.


\bibliography{iclr2023_conference_tinypaper}
\bibliographystyle{iclr2023_conference_tinypaper}

\appendix
\section{Appendix}

\subsection{Convergence of L-Tuning}
\label{sec:ablation_study}

\begin{figure}[htbp]
\centering
\includegraphics[width=0.65\linewidth]{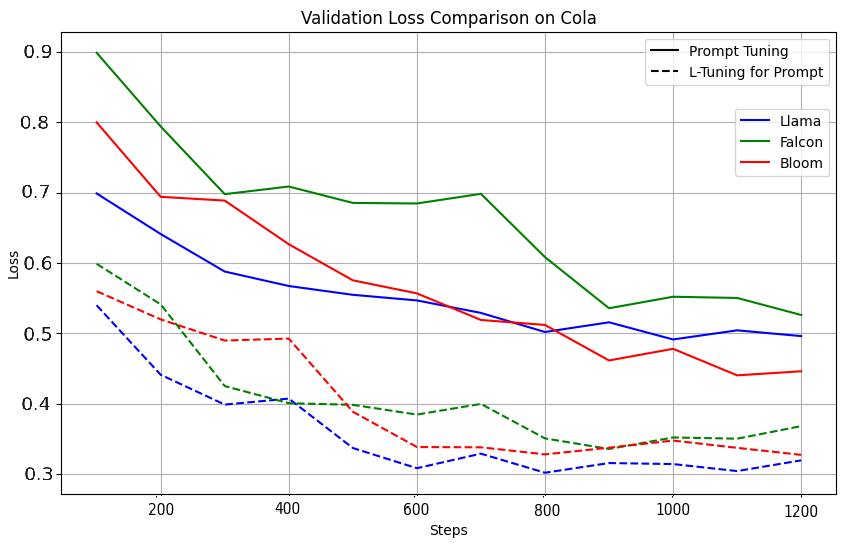} 
\caption{Validation Loss Comparison on the CoLA dataset across different steps for traditional Prompt Tuning (solid lines) and L-Tuning for Prompt (dashed lines) across models Llama (blue), Falcon (green), and Bloom (red).}
\label{fig:loss_comparison}
\end{figure}

In the ablation study presented, we focus on the convergence characteristics of L-Tuning in the context of prompt-based fine-tuning. Figure~\ref{fig:loss_comparison} illustrates the comparative validation loss between traditional Prompt Tuning and L-Tuning for Prompt across various training steps. The analysis involves three distinct pre-trained language models: Llama, Falcon, and Bloom.

The validation loss trajectories reveal a consistent pattern of faster convergence for L-Tuning for Prompt, in comparison to traditional Prompt Tuning. The prompt embedding strategy of L-Tuning, which directly leverages the semantic content of real label text to inform the label embeddings, provides the LLMs with a head start in understanding the association between text and labels. Consequently, L-Tuning facilitates a more expedient decline in validation loss, signifying more efficient learning dynamics.

This efficiency in convergence is hypothesized to be due to the direct usage of label semantics within the LM, allowing the model to capitalize on the pre-trained knowledge of label contexts. As a result, the LM under L-Tuning demonstrates an enhanced ability to correlate label information with the input text, minimizing the loss at a notably faster rate. Such an approach could lead to significant reductions in the required computational resources and time for model fine-tuning.

\subsection{Training Algorithm}

\textbf{L-Tuning for Prefix:} 
The L-Tuning for Prefix algorithm \ref{prefix_algo} applies a unique approach to fine-tuning a pre-trained LLM for classification tasks. This method maintains the LLM's parameters frozen while training only a select set of parameters associated with label inputs. Each label input is processed through the frozen LLM to generate a contextual representation. This representation is then pooled and transformed through a trainable self-attention mechanism and used alongside the text input for classification. The algorithm iteratively updates only the parameters of the self-attention pooling and the transformation function to minimize the classification loss.

\begin{algorithm}[h] 
\caption{L-Tuning for Prefix} \label{prefix_algo}
\begin{algorithmic}[1]
\State \textbf{Input:} Pre-trained LM $\mathcal{M}$ with parameters $\Theta$, frozen during training
\State \textbf{Input:} Label inputs $\{\mathbf{y}_i\}_{i=1}^N$, Text inputs $\{\mathbf{x}_i\}_{i=1}^N$
\State \textbf{Initialize:} Trainable parameters $\Phi$, $\Psi$ and $\zeta$ 

\For{each label input $\mathbf{y}_i$ and text input $\mathbf{x}_i$}
    \State $\mathbf{h}_i \gets \mathcal{M}_{\Theta_{\text{frozen}}}(\mathbf{y}_i)$ \Comment{Process label input}
    \State $\mathbf{p}_i \gets Z_{\Psi}(\mathcal{F}_{\Phi}(\mathbf{h}_i))$ \Comment{Self-attention pooling and transformation}
    \State $\mathbf{o}_i \gets \mathcal{C}_\zeta(\mathcal{M}_{\Theta_{\text{frozen}}}(\mathbf{x}_i, \mathbf{p}_i))$ \Comment{Classification head}
    \State Compute loss $\mathcal{L}(\mathbf{o}_i, \text{true label})$
    \State Update $\Phi$, $\Psi$ and $\zeta$ to minimize loss
\EndFor
\end{algorithmic}
\end{algorithm}

\textbf{L-Tuning for Prompt:} 
In contrast, the L-Tuning for Prompt algorithm \ref{pompt_algo} modifies the prompt tuning approach by generating embeddings for both label and text inputs, which are then concatenated and processed for classification. Here, the LLM's parameters are also kept frozen, and only a specific set of trainable parameters associated with the label embeddings are updated. This approach aims to capture more nuanced relationships within the data by transforming the label input into an effective embedding for classification, with the training process focusing on optimizing these label embeddings.

\begin{algorithm}[h]
\caption{L-Tuning for Prompt} \label{pompt_algo}
\begin{algorithmic}[1]
\State \textbf{Input:} Pre-trained LM $\mathcal{M}$ with parameters $\Theta$, frozen during training
\State \textbf{Input:} Label inputs $\{\mathbf{y}_i\}_{i=1}^N$, Text inputs $\{\mathbf{x}_i\}_{i=1}^N$
\State \textbf{Initialize:} Trainable parameter $\gamma$ and and $\zeta$

\For{each label input $\mathbf{y}_i$ and text input $\mathbf{x}_i$}
    \State $\mathbf{h}_i \gets \mathcal{M}_{\Theta_{\text{frozen}}}(\mathbf{y}_i)$ \Comment{Process label input}
    \State $\mathbf{e(y_i)} \gets \mathcal{G}_{\gamma}(\mathbf{h}_i)$ \Comment{Get label embedding}
    \State $\mathbf{e(x_i)} \gets \text{Embedding from frozen LM}(\mathbf{x}_i)$ \Comment{Get text embedding}
    \State $\mathbf{o}_i \gets \mathcal{C}(\mathcal{M}_{\Theta_{\text{frozen}}}(\mathbf{e(y_i)} \oplus \mathbf{e(x_i)}))$ \Comment{Concatenate embeddings and classify}
    \State Compute loss $\mathcal{L}(\mathbf{o}_i, \text{true label})$
    \State Update $\gamma$ and $\zeta$ to minimize loss
\EndFor
\end{algorithmic}
\end{algorithm}

\subsection{Training Methodology}
\label{sec:training_methodology}

For training our model,  we construct training batches that consist of both positive and negative examples. A positive example is a true pair \((\mathbf{x}_i,\mathbf{y}_j)\) where \( y_i = \mathbf{y}_i \), while a negative example is constructed by pairing \(\mathbf{x}_i\) with a false label \( \mathbf{y}_i \) where \( y_i \neq \mathbf{y}_i \).

Formally, each batch \( B \) is composed of \( m \) examples, where \( m/2 \) are positive examples drawn from \( \mathcal{D} \) and the remaining \( m/2 \) are negative examples. The negative examples are created by randomly assigning false labels to the text inputs, ensuring that the false label does not match the true label associated with the text. The batch can be represented as follows:
\begin{equation}
B = \{(\mathbf{x}_i,\mathbf{y}_i, 1)\}_{i=1}^{m/2} \cup \{(\mathbf{x}_i,\mathbf{y}_j, 0)\}_{i=m/2 + 1}^{m},
\end{equation}
The training objective is to minimize the binary cross-entropy loss \( \mathcal{L} \) across all batches, defined as:
\begin{equation}
\mathcal{L} = -\frac{1}{m} \sum_{(\mathbf{x}_i, \mathbf{y}, c_B) \in B} [c_B\log \mathbf{o}_B + (1 - c_B) \log (1 - \mathbf{o}_B)],
\end{equation}
Where \( c_B\)  and \(\mathbf{o}_B \) denote the class labels and model's predicted probability of $B$ batch respectively.

The model is updated iteratively to minimize \( \mathcal{L} \), improving its ability to discriminate between true and false text-label pairs. This binary classification setup trains the model to better understand the nuances of text-label relationships, which is essential for NLI tasks.

\subsection{Evaluation Procedure}
\label{sec:evaluation_procedure}

The evaluation of our model's classification performance is formulated within an NLI-inspired framework. Let us denote the set of all possible labels as \( \mathcal{Y} = \{y_1, y_2, \ldots, y_K\} \), where \( K \) is the number of unique labels. The predicted label \( \hat{y}_i \) for \(( \mathbf{x}_i, \mathbf{Y}) \) is the one that maximizes the model's score, formally defined as:
\begin{equation}
\hat{y}_i(\mathbf{x}_i) = \underset{y_k \in \mathcal{Y}}{\mathrm{argmax}} \; \mathbf{o}_i.
\end{equation}
The argmax operation selects the label with the highest score as the predicted label, mirroring the judgment process in NLI where the premise (text) is evaluated against multiple hypotheses (labels) to determine the most probable one.

The model's classification accuracy is then calculated as the proportion of text instances where the predicted label matches the true label \( y \):
\begin{equation}
\mathrm{Accuracy} = \frac{1}{N} \sum_{i=1}^{N} \mathbb{1}\{\hat{y}(\mathbf{x}_i) = \textbf{y}_i\},
\end{equation}
where \( N \) is the number of text instances in the evaluation dataset, and \( \mathbb{1} \) is the indicator function, which equals 1 when the predicted label matches the true label and 0 otherwise.

This metric evaluates the model's ability to correctly identify the label that best aligns with the semantic content of the text, providing a quantitative measure of its classification performance.

\subsection{Parameter Calculation}
\label{subsec:param_calc}

\textbf{L-Tuning for Prefix:}
In our prefix tuning approach, we implement a simplified self-attention pooling mechanism \(\mathcal{F}_{\Phi}\). This mechanism is designed to transform the last layer hidden representation \(\mathbf{h}\) from \(\mathbb{R}^{l \times d}\) to a pooled representation in \(\mathbb{R}^{d}\), where \(l\) is the sequence length of prefix and \(d\) is the hidden dimension \cite{safari2020self}.

The self-attention pooling applies a linear transformation, parameterized by \(\Phi\), to each \(d\)-dimensional vector in \(\mathbf{h}\), mapping it to a scalar, and producing \(l\) scalars in total. These scalars are then normalized through a softmax function to create attention weights, which are used to compute a weighted sum of the original \(l \times d\) matrix, resulting in a single \(l\)-dimensional vector. Since biases are not used, the number of trainable parameters in this linear transformation is \(d\).

Let \(m\) denote the number of layers in the LLM. The transformation function \(f_{\Psi}\) maps the pooled representation from \(\mathbb{R}^{l}\) to past key-value pairs for each layer, resulting in a representation of \(\mathbb{R}^{l \times m \times d}\). Given that there is a pair of past key values for each layer and assuming \(f_{\Psi}\) is a linear transformation without biases, the total number of trainable parameters for this transformation is \(2 \times l \times ( m \times d)\).

For the classification head, we used the pooling output; the trainable parameters amount to $2 \times d$

Therefore, the total number of trainable parameters for the prefix tuning in L-Tuning is:
\[ d + 2 \times l \times ( m \times d) + 2 \times d =  2 \times l \times m \times d + 3 \times d\]
This calculation indicates that our method employs \( 2 \times l\times m \times d + 3 \times d\) trainable parameters, distinguishing it from traditional prefix tuning methods, particularly in the additional parameters $d$ introduced by the self-attention pooling layer.

\textbf{L-Tuning for Prompt:}
In the case of L-Tuning for the prompt, our approach employs \(d^2\) trainable parameters in a linear transformation, \(\mathcal{G}_{\gamma}\), which converts a representation from \(\mathbb{R}^{l \times d}\) to \(\mathbb{R}^{l \times d}\). Additionally, there are \(2 \times d\) parameters for the classification head, totaling \(d(d + 2)\) trainable parameters. This contrasts with the traditional method of prompt tuning, which typically involves approximately \(d(l+k)\) parameters, where \(k\) is the total number of labels. The use of a square matrix in \(\mathcal{G}_{\gamma}\) allows for a more complex and nuanced transformation of the label embeddings, potentially capturing more intricate relationships within the data.

\end{document}

%% file: math_commands.tex
\usepackage{amsmath,amsfonts,bm}

\def\eqref#1{equation~\ref{#1}}

\def\1{\bm{1}}

\DeclareMathAlphabet{\mathsfit}{\encodingdefault}{\sfdefault}{m}{sl}
\SetMathAlphabet{\mathsfit}{bold}{\encodingdefault}{\sfdefault}{bx}{n}













%% file: iclr2023_conference_tinypaper.bbl
\begin{thebibliography}{15}
\providecommand{\natexlab}[1]{#1}
\providecommand{\url}[1]{\texttt{#1}}
\expandafter\ifx\csname urlstyle\endcsname\relax
  \providecommand{\doi}[1]{doi: #1}\else
  \providecommand{\doi}{doi: \begingroup \urlstyle{rm}\Url}\fi

\bibitem[Devlin et~al.(2019)Devlin, Chang, Lee, and Toutanova]{devlin-etal-2019-bert}
Jacob Devlin, Ming-Wei Chang, Kenton Lee, and Kristina Toutanova.
\newblock {BERT}: Pre-training of deep bidirectional transformers for language understanding.
\newblock In Jill Burstein, Christy Doran, and Thamar Solorio (eds.), \emph{Proceedings of the 2019 Conference of the North {A}merican Chapter of the Association for Computational Linguistics: Human Language Technologies, Volume 1 (Long and Short Papers)}, pp.\  4171--4186, Minneapolis, Minnesota, June 2019. Association for Computational Linguistics.
\newblock \doi{10.18653/v1/N19-1423}.
\newblock URL \url{https://aclanthology.org/N19-1423}.

\bibitem[Ge et~al.(2023)Ge, Hua, Ji, Tan, Xu, and Zhang]{ge2023openagi}
Yingqiang Ge, Wenyue Hua, Jianchao Ji, Juntao Tan, Shuyuan Xu, and Yongfeng Zhang.
\newblock Openagi: When llm meets domain experts.
\newblock \emph{arXiv preprint arXiv:2304.04370}, 2023.

\bibitem[Gu et~al.(2021)Gu, Han, Liu, and Huang]{gu2021ppt}
Yuxian Gu, Xu~Han, Zhiyuan Liu, and Minlie Huang.
\newblock Ppt: Pre-trained prompt tuning for few-shot learning.
\newblock \emph{arXiv preprint arXiv:2109.04332}, 2021.

\bibitem[Han et~al.(2022)Han, Zhao, Ding, Liu, and Sun]{han2022ptr}
Xu~Han, Weilin Zhao, Ning Ding, Zhiyuan Liu, and Maosong Sun.
\newblock Ptr: Prompt tuning with rules for text classification.
\newblock \emph{AI Open}, 3:\penalty0 182--192, 2022.

\bibitem[He et~al.(2021)He, Liu, Gao, and Chen]{he2021deberta}
Pengcheng He, Xiaodong Liu, Jianfeng Gao, and Weizhu Chen.
\newblock Deberta: Decoding-enhanced bert with disentangled attention, 2021.

\bibitem[Kowsher et~al.(2023)Kowsher, Sobuj, Prottasha, Arefin, and Morimoto]{kowsher2023contrastive}
Md~Kowsher, Md~Shohanur~Islam Sobuj, Nusrat~Jahan Prottasha, Mohammad~Shamsul Arefin, and Yasuhiko Morimoto.
\newblock Contrastive learning for universal zero-shot nli with cross-lingual sentence embeddings.
\newblock In \emph{Proceedings of the 3rd Workshop on Multi-lingual Representation Learning (MRL)}, pp.\  239--252, 2023.

\bibitem[Lester et~al.(2021)Lester, Al-Rfou, and Constant]{lester2021power}
Brian Lester, Rami Al-Rfou, and Noah Constant.
\newblock The power of scale for parameter-efficient prompt tuning.
\newblock \emph{arXiv preprint arXiv:2104.08691}, 2021.

\bibitem[Liu et~al.(2022)Liu, Ji, Fu, Tam, Du, Yang, and Tang]{liu2022p}
Xiao Liu, Kaixuan Ji, Yicheng Fu, Weng Tam, Zhengxiao Du, Zhilin Yang, and Jie Tang.
\newblock P-tuning: Prompt tuning can be comparable to fine-tuning across scales and tasks.
\newblock In \emph{Proceedings of the 60th Annual Meeting of the Association for Computational Linguistics (Volume 2: Short Papers)}, pp.\  61--68, 2022.

\bibitem[Liu et~al.(2019)Liu, Ott, Goyal, Du, Joshi, Chen, Levy, Lewis, Zettlemoyer, and Stoyanov]{liu2019roberta}
Yinhan Liu, Myle Ott, Naman Goyal, Jingfei Du, Mandar Joshi, Danqi Chen, Omer Levy, Mike Lewis, Luke Zettlemoyer, and Veselin Stoyanov.
\newblock Roberta: A robustly optimized bert pretraining approach, 2019.

\bibitem[Penedo et~al.(2023)Penedo, Malartic, Hesslow, Cojocaru, Cappelli, Alobeidli, Pannier, Almazrouei, and Launay]{penedo2023refinedweb}
Guilherme Penedo, Quentin Malartic, Daniel Hesslow, Ruxandra Cojocaru, Alessandro Cappelli, Hamza Alobeidli, Baptiste Pannier, Ebtesam Almazrouei, and Julien Launay.
\newblock The refinedweb dataset for falcon llm: Outperforming curated corpora with web data, and web data only, 2023.

\bibitem[Peng et~al.(2023)Peng, Wu, and Fang]{peng2023soft}
Zhiyuan Peng, Xuyang Wu, and Yi~Fang.
\newblock Soft prompt tuning for augmenting dense retrieval with large language models.
\newblock \emph{arXiv preprint arXiv:2307.08303}, 2023.

\bibitem[Safari et~al.(2020)Safari, India, and Hernando]{safari2020self}
Pooyan Safari, Miquel India, and Javier Hernando.
\newblock Self-attention encoding and pooling for speaker recognition.
\newblock \emph{arXiv preprint arXiv:2008.01077}, 2020.

\bibitem[Touvron et~al.(2023)Touvron, Martin, Stone, Albert, Almahairi, Babaei, Bashlykov, Batra, Bhargava, Bhosale, Bikel, Blecher, Ferrer, Chen, Cucurull, Esiobu, Fernandes, Fu, Fu, Fuller, Gao, Goswami, Goyal, Hartshorn, Hosseini, Hou, Inan, Kardas, Kerkez, Khabsa, Kloumann, Korenev, Koura, Lachaux, Lavril, Lee, Liskovich, Lu, Mao, Martinet, Mihaylov, Mishra, Molybog, Nie, Poulton, Reizenstein, Rungta, Saladi, Schelten, Silva, Smith, Subramanian, Tan, Tang, Taylor, Williams, Kuan, Xu, Yan, Zarov, Zhang, Fan, Kambadur, Narang, Rodriguez, Stojnic, Edunov, and Scialom]{touvron2023llama}
Hugo Touvron, Louis Martin, Kevin Stone, Peter Albert, Amjad Almahairi, Yasmine Babaei, Nikolay Bashlykov, Soumya Batra, Prajjwal Bhargava, Shruti Bhosale, Dan Bikel, Lukas Blecher, Cristian~Canton Ferrer, Moya Chen, Guillem Cucurull, David Esiobu, Jude Fernandes, Jeremy Fu, Wenyin Fu, Brian Fuller, Cynthia Gao, Vedanuj Goswami, Naman Goyal, Anthony Hartshorn, Saghar Hosseini, Rui Hou, Hakan Inan, Marcin Kardas, Viktor Kerkez, Madian Khabsa, Isabel Kloumann, Artem Korenev, Punit~Singh Koura, Marie-Anne Lachaux, Thibaut Lavril, Jenya Lee, Diana Liskovich, Yinghai Lu, Yuning Mao, Xavier Martinet, Todor Mihaylov, Pushkar Mishra, Igor Molybog, Yixin Nie, Andrew Poulton, Jeremy Reizenstein, Rashi Rungta, Kalyan Saladi, Alan Schelten, Ruan Silva, Eric~Michael Smith, Ranjan Subramanian, Xiaoqing~Ellen Tan, Binh Tang, Ross Taylor, Adina Williams, Jian~Xiang Kuan, Puxin Xu, Zheng Yan, Iliyan Zarov, Yuchen Zhang, Angela Fan, Melanie Kambadur, Sharan Narang, Aurelien Rodriguez, Robert Stojnic, Sergey Edunov, and Thomas
  Scialom.
\newblock Llama 2: Open foundation and fine-tuned chat models, 2023.

\bibitem[Wang et~al.(2019)Wang, Singh, Michael, Hill, Levy, and Bowman]{wang2019glue}
Alex Wang, Amanpreet Singh, Julian Michael, Felix Hill, Omer Levy, and Samuel~R. Bowman.
\newblock {GLUE}: A multi-task benchmark and analysis platform for natural language understanding.
\newblock 2019.
\newblock In the Proceedings of ICLR.

\bibitem[Workshop et~al.(2023)Workshop, :, Scao, Fan, Akiki, Pavlick, Ilić, Hesslow, Castagné, Luccioni, Yvon, Gallé, Tow, Rush, Biderman, Webson, Ammanamanchi, Wang, Sagot, Muennighoff, del Moral, Ruwase, Bawden, Bekman, McMillan-Major, Beltagy, Nguyen, Saulnier, Tan, Suarez, Sanh, Laurençon, Jernite, Launay, Mitchell, Raffel, Gokaslan, Simhi, Soroa, Aji, Alfassy, Rogers, Nitzav, Xu, Mou, Emezue, Klamm, Leong, van Strien, Adelani, Radev, Ponferrada, Levkovizh, Kim, Natan, Toni, Dupont, Kruszewski, Pistilli, Elsahar, Benyamina, Tran, Yu, Abdulmumin, Johnson, Gonzalez-Dios, de~la Rosa, Chim, Dodge, Zhu, Chang, Frohberg, Tobing, Bhattacharjee, Almubarak, Chen, Lo, Werra, Weber, Phan, allal, Tanguy, Dey, Muñoz, Masoud, Grandury, Šaško, Huang, Coavoux, Singh, Jiang, Vu, Jauhar, Ghaleb, Subramani, Kassner, Khamis, Nguyen, Espejel, de~Gibert, Villegas, Henderson, Colombo, Amuok, Lhoest, Harliman, Bommasani, López, Ribeiro, Osei, Pyysalo, Nagel, Bose, Muhammad, Sharma, Longpre, Nikpoor, Silberberg, Pai,
  Zink, Torrent, Schick, Thrush, Danchev, Nikoulina, Laippala, Lepercq, Prabhu, Alyafeai, Talat, Raja, Heinzerling, Si, Taşar, Salesky, Mielke, Lee, Sharma, Santilli, Chaffin, Stiegler, Datta, Szczechla, Chhablani, Wang, Pandey, Strobelt, Fries, Rozen, Gao, Sutawika, Bari, Al-shaibani, Manica, Nayak, Teehan, Albanie, Shen, Ben-David, Bach, Kim, Bers, Fevry, Neeraj, Thakker, Raunak, Tang, Yong, Sun, Brody, Uri, Tojarieh, Roberts, Chung, Tae, Phang, Press, Li, Narayanan, Bourfoune, Casper, Rasley, Ryabinin, Mishra, Zhang, Shoeybi, Peyrounette, Patry, Tazi, Sanseviero, von Platen, Cornette, Lavallée, Lacroix, Rajbhandari, Gandhi, Smith, Requena, Patil, Dettmers, Baruwa, Singh, Cheveleva, Ligozat, Subramonian, Névéol, Lovering, Garrette, Tunuguntla, Reiter, Taktasheva, Voloshina, Bogdanov, Winata, Schoelkopf, Kalo, Novikova, Forde, Clive, Kasai, Kawamura, Hazan, Carpuat, Clinciu, Kim, Cheng, Serikov, Antverg, van~der Wal, Zhang, Zhang, Gehrmann, Mirkin, Pais, Shavrina, Scialom, Yun, Limisiewicz, Rieser,
  Protasov, Mikhailov, Pruksachatkun, Belinkov, Bamberger, Kasner, Rueda, Pestana, Feizpour, Khan, Faranak, Santos, Hevia, Unldreaj, Aghagol, Abdollahi, Tammour, HajiHosseini, Behroozi, Ajibade, Saxena, Ferrandis, McDuff, Contractor, Lansky, David, Kiela, Nguyen, Tan, Baylor, Ozoani, Mirza, Ononiwu, Rezanejad, Jones, Bhattacharya, Solaiman, Sedenko, Nejadgholi, Passmore, Seltzer, Sanz, Dutra, Samagaio, Elbadri, Mieskes, Gerchick, Akinlolu, McKenna, Qiu, Ghauri, Burynok, Abrar, Rajani, Elkott, Fahmy, Samuel, An, Kromann, Hao, Alizadeh, Shubber, Wang, Roy, Viguier, Le, Oyebade, Le, Yang, Nguyen, Kashyap, Palasciano, Callahan, Shukla, Miranda-Escalada, Singh, Beilharz, Wang, Brito, Zhou, Jain, Xu, Fourrier, Periñán, Molano, Yu, Manjavacas, Barth, Fuhrimann, Altay, Bayrak, Burns, Vrabec, Bello, Dash, Kang, Giorgi, Golde, Posada, Sivaraman, Bulchandani, Liu, Shinzato, de~Bykhovetz, Takeuchi, Pàmies, Castillo, Nezhurina, Sänger, Samwald, Cullan, Weinberg, Wolf, Mihaljcic, Liu, Freidank, Kang, Seelam, Dahlberg,
  Broad, Muellner, Fung, Haller, Chandrasekhar, Eisenberg, Martin, Canalli, Su, Su, Cahyawijaya, Garda, Deshmukh, Mishra, Kiblawi, Ott, Sang-aroonsiri, Kumar, Schweter, Bharati, Laud, Gigant, Kainuma, Kusa, Labrak, Bajaj, Venkatraman, Xu, Xu, Xu, Tan, Xie, Ye, Bras, Belkada, and Wolf]{workshop2023bloom}
BigScience Workshop, :, Teven~Le Scao, Angela Fan, Christopher Akiki, Ellie Pavlick, Suzana Ilić, Daniel Hesslow, Roman Castagné, Alexandra~Sasha Luccioni, François Yvon, Matthias Gallé, Jonathan Tow, Alexander~M. Rush, Stella Biderman, Albert Webson, Pawan~Sasanka Ammanamanchi, Thomas Wang, Benoît Sagot, Niklas Muennighoff, Albert~Villanova del Moral, Olatunji Ruwase, Rachel Bawden, Stas Bekman, Angelina McMillan-Major, Iz~Beltagy, Huu Nguyen, Lucile Saulnier, Samson Tan, Pedro~Ortiz Suarez, Victor Sanh, Hugo Laurençon, Yacine Jernite, Julien Launay, Margaret Mitchell, Colin Raffel, Aaron Gokaslan, Adi Simhi, Aitor Soroa, Alham~Fikri Aji, Amit Alfassy, Anna Rogers, Ariel~Kreisberg Nitzav, Canwen Xu, Chenghao Mou, Chris Emezue, Christopher Klamm, Colin Leong, Daniel van Strien, David~Ifeoluwa Adelani, Dragomir Radev, Eduardo~González Ponferrada, Efrat Levkovizh, Ethan Kim, Eyal~Bar Natan, Francesco~De Toni, Gérard Dupont, Germán Kruszewski, Giada Pistilli, Hady Elsahar, Hamza Benyamina, Hieu Tran,
  Ian Yu, Idris Abdulmumin, Isaac Johnson, Itziar Gonzalez-Dios, Javier de~la Rosa, Jenny Chim, Jesse Dodge, Jian Zhu, Jonathan Chang, Jörg Frohberg, Joseph Tobing, Joydeep Bhattacharjee, Khalid Almubarak, Kimbo Chen, Kyle Lo, Leandro~Von Werra, Leon Weber, Long Phan, Loubna~Ben allal, Ludovic Tanguy, Manan Dey, Manuel~Romero Muñoz, Maraim Masoud, María Grandury, Mario Šaško, Max Huang, Maximin Coavoux, Mayank Singh, Mike Tian-Jian Jiang, Minh~Chien Vu, Mohammad~A. Jauhar, Mustafa Ghaleb, Nishant Subramani, Nora Kassner, Nurulaqilla Khamis, Olivier Nguyen, Omar Espejel, Ona de~Gibert, Paulo Villegas, Peter Henderson, Pierre Colombo, Priscilla Amuok, Quentin Lhoest, Rheza Harliman, Rishi Bommasani, Roberto~Luis López, Rui Ribeiro, Salomey Osei, Sampo Pyysalo, Sebastian Nagel, Shamik Bose, Shamsuddeen~Hassan Muhammad, Shanya Sharma, Shayne Longpre, Somaieh Nikpoor, Stanislav Silberberg, Suhas Pai, Sydney Zink, Tiago~Timponi Torrent, Timo Schick, Tristan Thrush, Valentin Danchev, Vassilina Nikoulina,
  Veronika Laippala, Violette Lepercq, Vrinda Prabhu, Zaid Alyafeai, Zeerak Talat, Arun Raja, Benjamin Heinzerling, Chenglei Si, Davut~Emre Taşar, Elizabeth Salesky, Sabrina~J. Mielke, Wilson~Y. Lee, Abheesht Sharma, Andrea Santilli, Antoine Chaffin, Arnaud Stiegler, Debajyoti Datta, Eliza Szczechla, Gunjan Chhablani, Han Wang, Harshit Pandey, Hendrik Strobelt, Jason~Alan Fries, Jos Rozen, Leo Gao, Lintang Sutawika, M~Saiful Bari, Maged~S. Al-shaibani, Matteo Manica, Nihal Nayak, Ryan Teehan, Samuel Albanie, Sheng Shen, Srulik Ben-David, Stephen~H. Bach, Taewoon Kim, Tali Bers, Thibault Fevry, Trishala Neeraj, Urmish Thakker, Vikas Raunak, Xiangru Tang, Zheng-Xin Yong, Zhiqing Sun, Shaked Brody, Yallow Uri, Hadar Tojarieh, Adam Roberts, Hyung~Won Chung, Jaesung Tae, Jason Phang, Ofir Press, Conglong Li, Deepak Narayanan, Hatim Bourfoune, Jared Casper, Jeff Rasley, Max Ryabinin, Mayank Mishra, Minjia Zhang, Mohammad Shoeybi, Myriam Peyrounette, Nicolas Patry, Nouamane Tazi, Omar Sanseviero, Patrick von
  Platen, Pierre Cornette, Pierre~François Lavallée, Rémi Lacroix, Samyam Rajbhandari, Sanchit Gandhi, Shaden Smith, Stéphane Requena, Suraj Patil, Tim Dettmers, Ahmed Baruwa, Amanpreet Singh, Anastasia Cheveleva, Anne-Laure Ligozat, Arjun Subramonian, Aurélie Névéol, Charles Lovering, Dan Garrette, Deepak Tunuguntla, Ehud Reiter, Ekaterina Taktasheva, Ekaterina Voloshina, Eli Bogdanov, Genta~Indra Winata, Hailey Schoelkopf, Jan-Christoph Kalo, Jekaterina Novikova, Jessica~Zosa Forde, Jordan Clive, Jungo Kasai, Ken Kawamura, Liam Hazan, Marine Carpuat, Miruna Clinciu, Najoung Kim, Newton Cheng, Oleg Serikov, Omer Antverg, Oskar van~der Wal, Rui Zhang, Ruochen Zhang, Sebastian Gehrmann, Shachar Mirkin, Shani Pais, Tatiana Shavrina, Thomas Scialom, Tian Yun, Tomasz Limisiewicz, Verena Rieser, Vitaly Protasov, Vladislav Mikhailov, Yada Pruksachatkun, Yonatan Belinkov, Zachary Bamberger, Zdeněk Kasner, Alice Rueda, Amanda Pestana, Amir Feizpour, Ammar Khan, Amy Faranak, Ana Santos, Anthony Hevia, Antigona
  Unldreaj, Arash Aghagol, Arezoo Abdollahi, Aycha Tammour, Azadeh HajiHosseini, Bahareh Behroozi, Benjamin Ajibade, Bharat Saxena, Carlos~Muñoz Ferrandis, Daniel McDuff, Danish Contractor, David Lansky, Davis David, Douwe Kiela, Duong~A. Nguyen, Edward Tan, Emi Baylor, Ezinwanne Ozoani, Fatima Mirza, Frankline Ononiwu, Habib Rezanejad, Hessie Jones, Indrani Bhattacharya, Irene Solaiman, Irina Sedenko, Isar Nejadgholi, Jesse Passmore, Josh Seltzer, Julio~Bonis Sanz, Livia Dutra, Mairon Samagaio, Maraim Elbadri, Margot Mieskes, Marissa Gerchick, Martha Akinlolu, Michael McKenna, Mike Qiu, Muhammed Ghauri, Mykola Burynok, Nafis Abrar, Nazneen Rajani, Nour Elkott, Nour Fahmy, Olanrewaju Samuel, Ran An, Rasmus Kromann, Ryan Hao, Samira Alizadeh, Sarmad Shubber, Silas Wang, Sourav Roy, Sylvain Viguier, Thanh Le, Tobi Oyebade, Trieu Le, Yoyo Yang, Zach Nguyen, Abhinav~Ramesh Kashyap, Alfredo Palasciano, Alison Callahan, Anima Shukla, Antonio Miranda-Escalada, Ayush Singh, Benjamin Beilharz, Bo~Wang, Caio Brito,
  Chenxi Zhou, Chirag Jain, Chuxin Xu, Clémentine Fourrier, Daniel~León Periñán, Daniel Molano, Dian Yu, Enrique Manjavacas, Fabio Barth, Florian Fuhrimann, Gabriel Altay, Giyaseddin Bayrak, Gully Burns, Helena~U. Vrabec, Imane Bello, Ishani Dash, Jihyun Kang, John Giorgi, Jonas Golde, Jose~David Posada, Karthik~Rangasai Sivaraman, Lokesh Bulchandani, Lu~Liu, Luisa Shinzato, Madeleine~Hahn de~Bykhovetz, Maiko Takeuchi, Marc Pàmies, Maria~A Castillo, Marianna Nezhurina, Mario Sänger, Matthias Samwald, Michael Cullan, Michael Weinberg, Michiel~De Wolf, Mina Mihaljcic, Minna Liu, Moritz Freidank, Myungsun Kang, Natasha Seelam, Nathan Dahlberg, Nicholas~Michio Broad, Nikolaus Muellner, Pascale Fung, Patrick Haller, Ramya Chandrasekhar, Renata Eisenberg, Robert Martin, Rodrigo Canalli, Rosaline Su, Ruisi Su, Samuel Cahyawijaya, Samuele Garda, Shlok~S Deshmukh, Shubhanshu Mishra, Sid Kiblawi, Simon Ott, Sinee Sang-aroonsiri, Srishti Kumar, Stefan Schweter, Sushil Bharati, Tanmay Laud, Théo Gigant, Tomoya
  Kainuma, Wojciech Kusa, Yanis Labrak, Yash~Shailesh Bajaj, Yash Venkatraman, Yifan Xu, Yingxin Xu, Yu~Xu, Zhe Tan, Zhongli Xie, Zifan Ye, Mathilde Bras, Younes Belkada, and Thomas Wolf.
\newblock Bloom: A 176b-parameter open-access multilingual language model, 2023.

\end{thebibliography}
